\documentclass[conference,usletter]{IEEEtran}
\usepackage{times,amsmath,amssymb}
\usepackage{slashbox}
\usepackage{graphicx}
\usepackage{xcolor}
\usepackage{algorithm2e}
\usepackage{algorithmicx}
\usepackage{setspace}
\usepackage[export]{adjustbox}
\usepackage{multirow}
\usepackage[noend]{algpseudocode}

\begin{document}

\title{Attention-based Neural Bag-of-Features Learning for Sequence Data}
\author{\IEEEauthorblockN{Dat Thanh Tran\IEEEauthorrefmark{1}, Nikolaos Passalis\IEEEauthorrefmark{2}, Anastasios Tefas\IEEEauthorrefmark{2}, Moncef Gabbouj\IEEEauthorrefmark{1}, Alexandros Iosifidis\IEEEauthorrefmark{3}}
\IEEEauthorblockA{\IEEEauthorrefmark{1}Department of Computing Sciences, Tampere University, Tampere, Finland\\
\IEEEauthorrefmark{2}Department of Informatics, Aristotle University of Thessaloniki, Thessaloniki, Greece\\
\IEEEauthorrefmark{3}Department of Engineering, Aarhus University, Aarhus, Denmark\\
thanh.tran@tuni.fi, passalis@csd.auth.gr, tefas@aiia.csd.auth.gr, \\ 
moncef.gabbouj@tuni.fi, alexandros.iosifidis@eng.au.dk}\\

}

\maketitle

\begin{abstract}
In this paper, we propose 2D-Attention (2DA), a generic attention formulation for sequence data, which acts as a complementary computation block that can detect and focus on relevant sources of information for the given learning objective. The proposed attention module is incorporated into the recently proposed Neural Bag of Feature (NBoF) model to enhance its learning capacity. Since 2DA acts as a plug-in layer, injecting it into different computation stages of the NBoF model results in different 2DA-NBoF architectures, each of which possesses a unique interpretation. We conducted extensive experiments in financial forecasting, audio analysis as well as medical diagnosis problems to benchmark the proposed formulations in comparison with existing methods, including the widely used Gated Recurrent Units. Our empirical analysis shows that the proposed attention formulations can not only improve performances of NBoF models but also make them resilient to noisy data. 
\end{abstract}

\section{Introduction}\label{introduction}
Learning problems in many fields involve sequence data such as time-series forecasting \cite{tran2017tensor, ntakaris2018benchmark}, audio analysis \cite{graves2013speech, stowell2015detection} or natural language processing \cite{cho2014learning, bahdanau2014neural}, all of which have been extensively studied. In many application scenarios, the observed sequence is highly non-stationary and noisy, which makes the task of modeling the underlying generating process more difficult. For example, in sound source separation in which the objective is to recover different unknown sources by filtering the observed mixtures, the existence of environmental noise is inherent and often complicates the separation process. Several mathematical techniques have been proposed to model the underlying data and noise distributions or to extract hand-crafted features, capturing certain desirable properties. In financial time-series analysis, representative examples include autoregressive (AR) and moving average (MA) \cite{slutzky1937summation} features, which were later extended with a differencing step to eliminate nonstationarity, known as autoregressive integrated moving average (ARIMA) \cite{tiao1981modeling}. Gaussian processes and Hidden Markov Model were popular mathematical frameworks in audio analysis. To ensure mathematical and computational tractability, these classical models are often formulated under many assumptions, which are sensitive to initialization and misaligned with real-world conditions, thus limiting their professional usage in practice. 

During the last decade, thanks to the development in stochastic optimization techniques and computing hardware, as well as the declining costs of data acquisition and storage, a data-driven approach based on deep neural networks and stochastic optimization has replaced the classical model-based approach and convex optimization. Nowadays, many of the state-of-the-art solutions for learning with sequence data are developed on the basis of neural networks. Notably, a class of neural network architecture called Recurrent Neural Networks (RNN), which is specifically designed to process variable-length sequences and to capture sequential patterns, has become the main workforce in different application domains. Another dedicated neural formulation for sequence data is the bilinear structures \cite{tran2017tensor, tran2017multilinear, tran2018temporal}, which were proposed to separately capture the dependencies along the temporal and spatial dimension in financial time-series. Even existing neural architectures, which were originally proposed for visual inputs such as Convolutional Neural Network (CNN) \cite{lecun1998gradient} and Neural Bag-of-Features (NBoF) \cite{passalis2017neural}, have shown competitive performances in tackling sequence data compared to dedicated statistical models \cite{zhao2017convolutional, passalis2017time}. The advantage of neural formulations over statistical learning and traditional hand-crafted features lies in the fact that fewer assumptions are made, and data is leveraged to automatically identify and extract task-relevant features in an end-to-end fashion. 

Bag-of-Features (BoF) model \cite{lazebnik2006beyond} was originally proposed to build histogram representations from images. Later, it was shown that BoF could be successfully applied to extract high-level representations for other data modalities such as video and audio \cite{jiang2007towards, riley2008text, iosifidis2012multidimensional, iosifidis2014discriminant}. Learning BoF representations consists of two steps: \textit{dictionary learning} and \textit{feature quantization and encoding}. In the dictionary learning step, each object is first represented by a set of low-level features, which could be, for example, a collection of local descriptors like SIFT \cite{lowe1999object} for image object or word-level vector-encodings for a sentence object. These features are then used to generate a compact dictionary (codebook) comprising of the most representative features, also known as \textit{codewords}. In the second step, the histogram representation of each object is extracted by quantizing its low-level features using the codebook. 

Recently, Neural Bag-of-Features (NBoF) \cite{passalis2017neural}, a neural network generalization of the BoF model, has been proposed. Similar to its predecessor, NBoF can generate a fixed-size histogram vector from variable-size inputs. This neural network generalization works as a feature extraction layer, which can be combined and optimized jointly with other neural network layers to tackle both unsupervised and supervised objectives via stochastic optimization. Since the dictionary learning step in NBoF is updated in conjunction with other layers towards the end goal of optimizing an objective function, histogram vectors synthesized by NBoF are more representative than those produced by BoF in different learning scenarios such as visual recognition, information retrieval, and financial forecasting \cite{passalis2016entropy, passalis2017neural, passalis2017time}. 

While the NBoF model works well in different learning problems, the current formulations still possess some limitations. In the aggregation step, all of the quantized features are simply averaged to form the histogram vector. For sequence data, this implies that the model only allows equal contributions of the quantized features coming from different time steps to form the output representation. Similarly, the quantization results produced by each codeword are considered equally important for every sequence in the training set. These properties limit the dictionary learning, quantization, and encoding process to fully take advantage of the data-driven approach. 

To incorporate a higher degree of flexibility into the NBoF model, a weighing mechanism on a sequence level is desirable. That is, for each individual sequence, the model has the flexibility to perform a weighted sum of quantized outputs in the aggregation step, with the coefficients being adaptively changed with respect to the input sequence, or to select/discard irrelevant codewords, given the input sequence. In neural network literature, this is often achieved by having some attention mechanisms \cite{xu2015show, mnih2014recurrent, tran2018temporal}. The idea of attention is inspired by the phenomenon observed in the human visual cortex that visual stimuli from multiple objects actively compete for neural encoding. 

Although various attention mechanisms have been proposed for existing neural network architectures such as CNN \cite{xu2015show, laakom2019bag}, LSTM \cite{mnih2014recurrent, qin2017dual} or Bilinear structure \cite{tran2018temporal}, there is yet any formulation for the NBoF model when learning with sequence data. To have a generic attention mechanism that can be applied in a plug-and-play manner, in this work, we propose 2D-Attention (2DA), a neural network module that promotes competitions among different rows or columns in the input matrix and only (soft) selects those which win for attention. We will then demonstrate that by injecting 2DA into NBoF, we can overcome those limitations mentioned previously. The contributions of our work can be summarized as follows:
\begin{itemize}
\item We propose a new type of attention formulation for matrix data, which is dubbed as 2DA. The proposed layer acts as a complementary computation block, which is capable of identifying relevant sources of information to perform selective masking on the given input matrix.

\item We incorporate 2DA into different stages of the Neural Bag-of-Features (NBoF) model, creating various 2DA-NBoF extensions that can enhance the feature quantization or histogram accumulation step in the NBoF model. Extensive experiments were conducted in three different application domains: financial forecasting, audio analysis, and medical diagnosis, which demonstrate the effectiveness of our attention module in improving the NBoF model. In cases of noisy input, a variant of 2DA-NBoF shows resilience to noises by filtering out the noisy source of information before the feature quantization step. 
\end{itemize}

The remainder of the paper is organized as follows: in Section \ref{related-work}, we review the NBoF model and its extensions for time-series data, as well as previously proposed attention mechanisms in the neural network literature. In Section \ref{proposed-method}, we first present the proposed attention module 2DA and its interpretation. Several extensions of the NBoF model that incorporates 2DA are then presented. In Section \ref{experiments}, we provide details of our experiment protocols and quantitative analysis. Section \ref{conclusions} concludes our work with possible future research directions. 

\section{Related Work}\label{related-work}

The NBoF model \cite{passalis2017neural} consists of two components: a quantization layer and an accumulation layer. Each quantization neuron in the quantization layer performs like a codeword, which can be updated via BackPropagation algorithm. In the original formulation \cite{passalis2017neural}, the Radial Basis Function (RBF) layer was used for feature quantization. Recently, it has been shown that the hyperbolic kernel is also effective for the feature quantization step \cite{passalis2019temporal}. Here we describe the original formulation with RBF layer.  

Let $K$ be the number of neurons (codewords) in the RBF layer and $\mathbf{v}_k \in \mathbb{R}^D$ be the $k$-th codeword. In addition, the shape of the Gaussian function modeled by each neuron can be adjusted via parameter $\mathbf{w}_k$. Let us denote the sequence of $N$ features as $\mathbf{X} = [\mathbf{x}_1, \dots, \mathbf{x}_N] \in \mathbb{R}^{D \times N}$ with $\mathbf{x}_n \in \mathbb{R}^D, n=1, \dots, N$. The output of the $k$-th RBF neuron given the input feature $\mathbf{x}_n$ is the following:

\begin{equation}\label{eq1}
\phi_{n, k} = \frac{\textrm{exp}\big(-\|(\mathbf{x}_n - \mathbf{v_k})\odot \mathbf{w}_k \|_2\big)}
{\sum_{m=1}^{K}\textrm{exp}\big(-\|(\mathbf{x}_n - \mathbf{v}_m) \odot \mathbf{w}_m \|_2 \big)}
\end{equation}
where $\odot$ denotes element-wise product and $\mathbf{w}_k \in \mathbb{R}^D$ is the learnable weight vector that enables the shape of Gaussian kernel associated with the $k$-th RBF neuron to change.

As the sequence $\mathbf{X}$ goes through the quantization layer, each feature $\mathbf{x}_n$ is quantized as $\boldsymbol{\phi}_n = [\phi_{n, 1}, \dots, \phi_{n, K}]^T \in \mathbb{R}^K$, producing a sequence of quantized features $\boldsymbol{\Phi} = [\boldsymbol{\phi}_1, \dots, \boldsymbol{\phi}_N] \in \mathbb{R}^{K\times N}$. The accumulation layer aggregates the information in $\boldsymbol{\Phi}$ by calculating the averaged quantized feature:

\begin{equation}\label{eq2}
\mathbf{y} = \frac{1}{N} \sum_{n=1}^{N} \boldsymbol{\phi}_n
\end{equation}

There have been few extensions of NBoF model for sequence data. For example, Temporal Neural Bag-of-Features (TNBoF) model with different specialized codebooks has been proposed in \cite{passalis2018temporal} to capture both short-term and long-term temporal information in financial time-series. In \cite{passalis2019temporal}, the authors derived the logistic formulation of the NBoF model using the hyperbolic kernel instead of the RBF kernel for the quantization step and proposed an adaptive scaling mechanism which showed significant improvements in training stability and performance of the NBoF networks. 

While the attention mechanism was biologically inspired from the perspective of visual processing, this technique has also inspired and advanced several works in sequence data analysis, notably in sequence-to-sequence learning tasks. The first attention formulation applied to sequence data was proposed in \cite{bahdanau2014neural} for tackling machine translation tasks. In this formulation, the authors proposed to construct the context vectors in Sequence-to-sequence Recurrent Neural Network model by selectively combining some hidden states, rather than using the last hidden state as the context vector. The selection coefficients, also known as attention weights, are computed adaptively based on the given input sequence, and updated jointly with other parameters during stochastic optimization. 

The successful application of attention mechanism in machine translation tasks has led to the emergence of other attention formulations, which are designed to capture different types of salient information in sequence data. For example, in \cite{cinar2017position}, the authors proposed a formulation that can detect pseudo-periods in certain types of time-series, such as energy consumption or meteorology data. To predict the future stock index, a dual-stage attention mechanism was proposed in \cite{qin2017dual} for RNN to actively select relevant exogenous series and temporal instances. Similarly, to highlight and focus on important temporal events in Limit Order Book, the authors in \cite{tran2018temporal} proposed a method to calculate attention masks for bilinear networks. Although an attention formulation has been proposed for the convolutional NBoF model in \cite{laakom2019bag} to estimate the true color of images capturing by different devices, this formulation only works with image data. To the best of our knowledge, there has been no attention formulation for the NBoF model to tackle sequence data.

\section{Proposed Methods}\label{proposed-method}

In this Section, we will first present 2D-Attention (2DA), our proposed attention calculation for matrix data. Then, we will show how 2DA can be used to address different limitations of the NBoF model as described in Section \ref{introduction}. Throughout the paper, we denote scalar values by either lower-case or upper-case characters $(a,b,A,B,\dots)$, vectors by lower-case bold-face characters $(\mathbf{x}, \mathbf{y}, \dots)$, matrices by upper-case bold-face characters $(\mathbf{X}, \mathbf{Y},\dots )$, and mathematical functions by calligraphy characters $\mathcal{F}, \mathcal{G}, \dots$. In addition, we use $x_{mn}$ to denote the element at position $(m, n)$ in a matrix $\mathbf{X}$.  

\begin{figure*}[]
	\centering
	\includegraphics[width=0.99\textwidth]{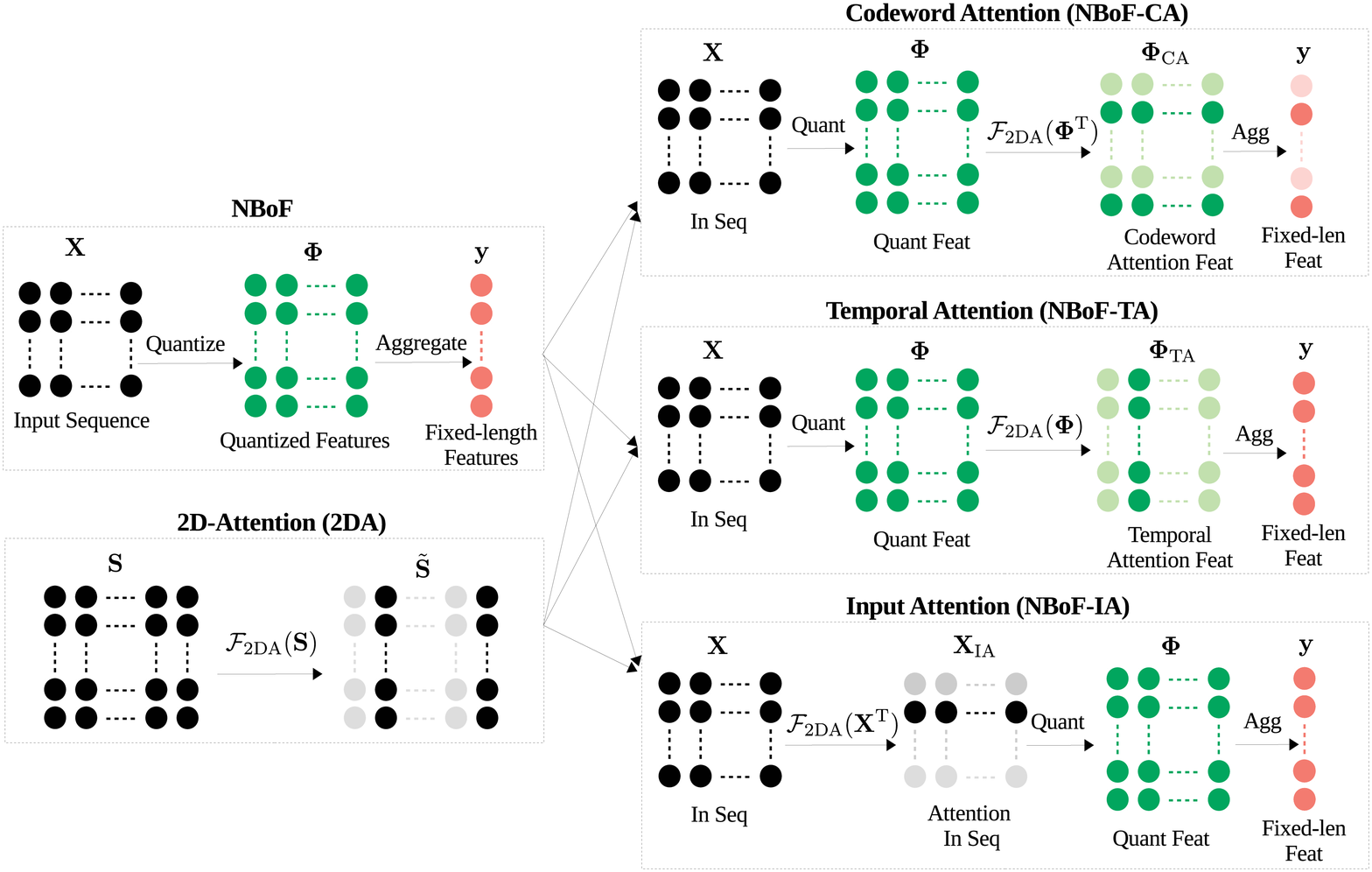}
	\caption{Illustration of the proposed attention formulation (2DA) and different attention-based NBoF models}\label{figure1}
\end{figure*}

\subsection{2D-Attention}\label{attention}

A matrix $\mathbf{S} \in \mathbb{R}^{M \times N}$ is a second-order tensor which has two modes, with $M$ and $N$ are the dimensions of the first and second mode, respectively. The matrix representation provides a natural way to represent a signal with two different sources of information. For example, a multivariate time-series is represented by a matrix with one mode representing the temporal dimension, and the other mode represents different sources that generate individual series. 

The general idea of attention mechanism is to highlight important elements in the data while discarding irrelevant ones. For data represented as a matrix $\mathbf{S}$, rather than considering each element in $\mathbf{S}$ individually, we would like to actively select certain columns or rows of $\mathbf{S}$ while discarding the others. This is because columns or rows of $\mathbf{S}$ usually form coherent sub-groups of the data. For example, discarding some temporal events or some individual series in a multivariate series corresponds to removing some rows or columns, depending on the orientation of the matrix.  

To adaptively determine and focus on different columns or rows of a matrix, we propose 2D-Attention (2DA), with the functional form denoted by $\mathcal{F}_{\mathrm{2DA}}$. This function takes a matrix $\mathbf{S} \in \mathbb{R}^{M \times N}$ as the input, and returns $\tilde{\mathbf{S}} \in \mathbb{R}^{M \times N}$ as the output. That is:

\begin{equation}\label{eq3}
\tilde{\mathbf{S}} = \mathcal{F}_{\mathrm{2DA}}(\mathbf{S})
\end{equation}

$\tilde{\mathbf{S}}$ can be considered as a filtered version of $\mathbf{S}$, where irrelevant columns of $\mathbf{S}$ with respect to the learning problem are zeroed out. Here we should note that $\mathcal{F}_{\mathrm{2DA}}$ performs adaptive attention with respect to columns of $\mathbf{S}$. To focus on different rows of $\mathbf{S}$, we can simply apply $\mathcal{F}_{\mathrm{2DA}}$ to the transpose of $\mathbf{S}$. 

The selection or rejection of the columns of $\mathbf{S}$ is conducted via element-wise matrix multiplications as follows:

\begin{equation}\label{eq4}
\tilde{\mathbf{S}} = \tau (\mathbf{S} \odot \mathbf{A}) + (1-\tau) \mathbf{S}
\end{equation}
where $\mathbf{A} \in \mathbb{R}^{M \times N}$ denotes the attention mask with values in the range $[0, 1]$. Each column in $\mathbf{A}$ encodes the importance of the corresponding column in $\mathbf{S}$. That is, the attention mask contains values that are close to $1$ corresponding to those columns in $\mathbf{S}$ that contain important information for the downstream learning task and vice versa. In Eq.(\ref{eq4}), parameter $\tau \in \mathbb{R}$, which is jointly optimized with other parameters, is used to allow flexible control of the attention mechanism: when $\mathbf{S}$ contains redundant or noisy information in its columns, the effect of attention mask $\mathbf{A}$ is enabled by pushing $\tau$ close to $1$; on the other hand, when every column of $\mathbf{S}$ is necessary, i.e., there is no need for attention, pushing $\tau$ close to $0$ will disable the effect of $\mathbf{A}$. The necessity of attention is thus automatically determined by optimizing $\tau$ with respect to a given problem. 

To calculate the attention mask $\mathbf{A}$, the proposed 2DA method learns to measure the relative importance between columns of $\mathbf{S}$ via a specially designed weight matrix $\mathbf{W}  \in \mathbb{R}^{N \times N}$: all elements of $\mathbf{W}$ are learnable, i.e., they are updated during stochastic optimization, except the diagonal elements, which are fixed to $1/N$. The attention mask is calculated as follows:

\begin{equation}\label{eq5}
\begin{split}
\mathbf{A} &= \mathcal{G}(\mathbf{Z}) \\
\mathbf{Z} &= \mathbf{S} \mathbf{W}
\end{split}
\end{equation}
where $\mathcal{G}(\mathbf{Z})$ denotes the soft-max function that is applied to every row of $\mathbf{Z}$. That is, every element of $\mathbf{A}$ is non-negative, and each row of $\mathbf
A$ sums up to $1$. Similar to other attention formulations \cite{qin2017dual, bahdanau2014neural, xu2015show}, we use soft-max normalization to promote competitions between different columns of $\mathbf{Z}$. 

As mentioned previously, the weight matrix $\mathbf{W}$ is used to measure the relative importance between columns of $\mathbf{S}$, which is encoded in $\mathbf{Z}$, and thus $\mathbf{A}$. In order to see this, let us denote by $\mathbf{s}_n \in \mathbb{R}^{M}$ and $\mathbf{z}_n \in \mathbb{R}^{M}$ the $n$-th column of $\mathbf{S}$ and $\mathbf{Z}$, respectively. Since $\mathbf{Z} = \mathbf{S} \mathbf{W}$, the $n$-th column of $\mathbf{Z}$, i.e., $\mathbf{z}_n$, is calculated as the weighted combination of $N$ columns of $\mathbf{S}$, with the weight of the $n$-th column always equal to $1/N$ since the diagonal elements of $\mathbf{W}$ are fixed to $1/N$. In this way, element $z_{mn}$ (in $\mathbf{Z}$) encodes the relative importance of $s_{mn}$ (in $\mathbf{S}$) with respect to other $s_{mk}$, for $k \neq n$.  

\subsection{Attention-based Neural Bag-of-Features}\label{nbof-attention}
In this subsection, we will show how the proposed attention module 2DA can be used to address different limitations of the NBoF model described in Section \ref{introduction}.

\textbf{Codeword Attention}: in the NBoF model, quantization results produced by each quantization neuron (codeword) are considered equally important for every input sequence. This property limits the feature quantization step to fully take advantage of the data-driven approach. In order to overcome this limitation, the proposed 2DA block can be applied to the quantized features to highlight or discard the outputs of certain quantization neurons. By doing so, the NBoF model is explicitly encouraged to learn a subset of specialized codewords for a given input pattern.

Particularly, given the quantized features denoted by $\mathbf{\Phi} \in \mathbb{R}^{K\times N}$ as described in Section \ref{related-work}, we propose to apply attention to the rows of $\mathbf{\Phi}$ because the first mode of $\mathbf{\Phi}$ with dimension $K$ denotes the number of quantization neurons or codewords. Since 2DA operates on the columns of the input matrix, the attention-based quantized features is calculated as follows: 

\begin{equation}\label{eq6}
\mathbf{\Phi}_{\textrm{CA}} = \mathcal{F}_{\mathrm{2DA}} (\mathbf{\Phi}^{\textrm{T}})
\end{equation}
where $\mathbf{\Phi}^{\textrm{T}}$ denotes the transpose of $\mathbf{\Phi}$. 

\textbf{Temporal Attention}: another limitation of the NBoF model lies in the aggregation step. In order to produce a fixed-length representation of the input sequence, the aggregation step in the NBoF model simply computes the mean of quantized features along the temporal mode. In this way, the NBoF model only allows equal contributions of all quantized features, disregarding the temporal information. In fact, the idea of giving different weights to different time instances has been adopted in previous works under different formulations \cite{tran2018temporal, qin2017dual}. Using our proposed 2DA formulation, it is straightforward to enable the NBoF model to attend to salient temporal information as follows:

\begin{equation}\label{eq7}
\mathbf{\Phi}_{\textrm{TA}} = \mathcal{F}_{\textrm{2DA}} (\mathbf{\Phi})
\end{equation}

Since each column of $\mathbf{\Phi}$ contains quantized features of each time step, to obtain temporal attention-based features $\mathbf{\Phi}_{\textrm{TA}}$ we simply apply $\mathcal{F}_{\textrm{2DA}}$ to $\mathbf{\Phi}$ as in Eq. (\ref{eq7}). $\mathbf{\Phi}_{\textrm{TA}}$ is then averaged along the second dimension to produce the fixed-length representation of the input sequence. Although we still perform averaging in the aggregation step, the fixed-length representation is no longer the average of the quantized features, but a weighted average. This is because each time instance (column) in $\mathbf{\Phi}_{\textrm{TA}}$ has been scaled by different factors via the attention mechanism.

\textbf{Input Attention}: noisy data is an inherent problem in many real-world applications. Noises might surface during the data acquisition process, such as ambient noise in audio signals or power line interference and motion artifacts in Electrocardiogram signals. In other scenarios, noises are inherent in the problem formulation since the relevance between the input sources and the targets might be unclear. For example, in stock prediction, it is intuitive to use related stocks' data, e.g., those coming from the same market sector, as the input to construct forecasting models, although some of them might be irrelevant to the movement of the target stock. 

The proposed attention mechanism can also be used to filter out potential noisy series in a multivariate series as follows:
\begin{equation}\label{eq8}
\mathbf{X}_{\textrm{IA}} = \mathcal{F}_{\textrm{2DA}}(\mathbf{X}^{\textrm{T}})
\end{equation}
where $\mathbf{X} \in \mathbb{R}^{D\times N}$ denotes an input sequence of $N$ steps of the NBoF model as specified in Section \ref{related-work}. Since we would like to apply attention over the individual series (rows of $\mathbf{X}$), $\mathcal{F}_{\textrm{2DA}}$ is applied to the transpose of $\mathbf{X}$. 

The proposed attention variants of the NBoF model are illustrated in Figure \ref{figure1}. 

\section{Experiments}\label{experiments}

In this section, we provide detailed descriptions and results of our empirical analysis, which demonstrate the advantages of attention-based NBoF models proposed in Section \ref{proposed-method}. Experiments were conducted in different types of sequence data, namely financial time-series in stock movement prediction problem, Electrocardiogram (ECG) and Phonocardiogram (PCG) in heart anomaly detection problems, and audio recording in music genre recognition and acoustic scene classification problems. 

The experiments were conducted with the recently proposed logistic formulation of the NBoF model \cite{passalis2019temporal}, i.e., the hyperbolic kernel was used in the quantization layer. In addition, we also experimented with the temporal variant of the NBoF model as proposed in \cite{passalis2018temporal} with a long-term and a short-term codebook. This variant is denoted as TNBoF. The codebook attention, temporal attention, and input attention when applied to the NBoF model are denoted as NBoF-CA, NBoF-TA, and NBoF-IA, respectively. The corresponding attention variants for the TNBoF model are denoted as TNBoF-CA, TNBoF-TA, and TNBoF-IA. In addition to the NBoF and TNBoF models serving as the baseline models, we also evaluated RNN models using Gated Recurrent Units (GRU) \cite{cho2014learning}.  

\subsection{Financial Forecasting Experiments}\label{finance}

Although extensively studied over the last decades, financial forecasting still remains as the most challenging tasks among time-series predictions \cite{sezer2020financial}. This is due to the complex dynamics of the financial markets, which make the observed data highly stationary and noisy. For this reason, we selected the stock movement prediction problem in FI2010 dataset \cite{ntakaris2018benchmark} as a representative problem in time-series forecasting. FI2010 is the largest publicly available Limit Order Book (LOB) dataset, which contains approximately $4.5$ million order events. The limit orders came from $5$ Finnish stocks traded in Helsinki Exchange (operated by NASDAQ Nordic) over $10$ business days. At each time instance, the dataset provides information (the prices and volumes) of the top $10$ levels, leading to a $40$-dimensional vector representation. 

The FI2010 dataset is used to investigate the problem of mid-price movement prediction in the next $H=\{10, 20, 50\}$ order events. The mid-price at a given time instance is the average between the best buy and best sell prices. This quantity is a virtual price since no trade can happen at this particular price at the given time instance. The movement of mid-price (stationary, increasing, decreasing) reflects the dynamic of the LOB and the market, thus plays an important role in financial analysis. The dataset provides the movement labels, given the future horizon $H=\{10, 20, 50\}$. Details regarding the FI2010 dataset and LOB can be found in \cite{ntakaris2018benchmark}. 

We followed the same experimental setup proposed in \cite{tran2018temporal}, which used the first $7$ days for training the models and the last $3$ days to test the performances. Due to the imbalanced nature of the problem, we reported averaged F1 score per movement as the main performance metric, similar to prior experiments \cite{tran2017tensor, tran2018temporal}. Detailed information about the training hyper-parameters and the network architectures is provided in the Appendix.

\begin{table}[t!]
	\begin{center}
		\caption{Results on the FI2010 dataset. The second column shows performances (averaged F1 in \%) of all models without using any convolution layer as preprocessing layers. The third column shows the corresponding performances (average F1 in \%)  when additional convolution layers were used as preprocessing layers. The best results in each column are highlighted in bold-face}\label{t1}
		\resizebox{0.83\linewidth}{!}{
			\begin{tabular}{|c|c|c|}
				\multicolumn{3}{c}{} \\ \hline
				\textbf{Models}        & \textbf{without conv}  & \textbf{with conv}               \\ \hline \hline
				\multicolumn{3}{|c|}{\textit{Prediction Horizon $H=10$}} \\ \hline \hline
		
				GRU \cite{cho2014learning}	  & $\textbf{60.92}\scriptstyle\pm 00.09$ & $62.21\scriptstyle\pm 00.30$ \\ \hline \hline
				NBoF \cite{passalis2019temporal}	 & $33.07\scriptstyle\pm 00.66$ & $66.34\scriptstyle\pm 00.60$ \\ \hline
				NBoF-CA (our)	 & $40.81\scriptstyle\pm 00.05$  & $67.56\scriptstyle\pm 00.02$ \\ \hline
				NBoF-TA (our) 	  & $40.83\scriptstyle\pm 00.21$ & $\textbf{67.98}\scriptstyle\pm 00.09$ \\ \hline \hline
				TNBoF \cite{passalis2018temporal}	 & $36.66\scriptstyle\pm 00.51$ & $66.74\scriptstyle\pm 00.36$ \\ \hline
				TNBoF-CA (our)	 & $45.61\scriptstyle\pm 00.16$ & $67.76\scriptstyle\pm 00.05$ \\ \hline
				TNBoF-TA (our)	 & $45.97\scriptstyle\pm 00.15$ & $67.88\scriptstyle\pm 00.13$ \\ \hline \hline
				\multicolumn{3}{|c|}{\textit{Prediction Horizon $H=20$}} \\ \hline \hline
				GRU \cite{cho2014learning}	 & $\textbf{51.61}\scriptstyle\pm 00.25$ & $53.83\scriptstyle\pm 00.14$ \\ \hline \hline 
				NBoF \cite{passalis2019temporal}	 & $38.06\scriptstyle\pm 00.53$ & $58.85\scriptstyle\pm 00.05$ \\ \hline
				NBoF-CA (our)		& $40.08\scriptstyle\pm 00.07$ & $59.31\scriptstyle\pm 00.44$ \\ \hline
				NBoF-TA (our)  & $40.34\scriptstyle\pm 00.06$ & $\textbf{60.10}\scriptstyle\pm 00.03$ \\ \hline \hline
				TNBoF \cite{passalis2018temporal}	 & $38.67\scriptstyle\pm 00.50$ & $59.61\scriptstyle\pm 00.48$\\ \hline
				TNBoF-CA (our)		& $43.06\scriptstyle\pm 00.03$ & $59.73\scriptstyle\pm 00.19$ \\ \hline
				TNBoF-TA (our)	 & $43.50\scriptstyle\pm 00.15$ & $60.04\scriptstyle\pm 00.24$ \\ \hline \hline
				\multicolumn{3}{|c|}{\textit{Prediction Horizon $H=50$}} \\ \hline \hline 
				GRU \cite{cho2014learning}		 & $\textbf{63.13}\scriptstyle\pm 00.19$ & $65.93\scriptstyle\pm 00.03$ \\ \hline \hline 
				NBoF \cite{passalis2019temporal}	 & $48.25\scriptstyle\pm 00.25$ & $68.84\scriptstyle\pm 02.29$ \\ \hline
				NBoF-CA (our)	 & $49.34\scriptstyle\pm 00.17$ & $73.25\scriptstyle\pm 00.27$ \\ \hline
				NBoF-TA	(our)  & $49.21\scriptstyle\pm 00.16$ & $73.02\scriptstyle\pm 00.04$ \\ \hline \hline
				TNBoF \cite{passalis2018temporal}	 & $54.06\scriptstyle\pm 00.14$ & $69.27\scriptstyle\pm 01.09$ \\ \hline 
				TNBoF-CA (our) 	& $57.15\scriptstyle\pm 00.21$ & $\textbf{73.77}\scriptstyle\pm 00.37$ \\ \hline
				TNBoF-TA (our) 	& $57.41\scriptstyle\pm 00.06$ & $73.40\scriptstyle\pm 00.08$ \\ \hline 
				
			\end{tabular}
		}
	\end{center}
\end{table}
 
The experiment results for FI2010 are shown in Table \ref{t1}. In the second column of Table \ref{t1}, we list the performances of all models without using any convolution layers as the preprocessing layers. That is, the results in the second column of Table \ref{t1} are produced by architectures consisting of only the layer of interests (such as GRU, NBoF, and so on), plus the fully connected layers for generating predictions. In this setting, the GRU models outperform all variants of the NBoF model. This is expected since the NBoF model, by construction, is not designed to capture local features and long-term dependency in the input sequence. We can easily observe that this limitation can be partially overcome with the TNBoF variant, which uses two separate codebooks to capture the short-term and long-term dependency. By applying our proposed attention mechanism, performances of both the NBoF and TNBoF models are further boosted.  

The third column of Table \ref{t1} shows the performances of all models when using two additional convolution layers as the local feature extractor, prior to applying the layer of interest. It is clear that all of the models benefit from the additional convolution layers, especially the NBoF model and its variants. In this setting, the GRU models no longer dominate the family of NBoF models. In fact, the GRU models become the worst-performing ones in the third column of Table \ref{t1}. Furthermore, both codebook attention (NBoF-CA, TNBoF-CA) and temporal attention (NBoF-TA, TNBoF-TA) consistently enhance the baselines' performances, making attention-based models the best-performing ones. 

Here we should note that although the baseline models (NBoF, TNBoF) use the adaptive scaling step proposed in \cite{passalis2019temporal} to improve training stability, we did not employ this step in attention-based models. The reason stems from the fact that adaptive scaling introduces additional degrees of freedom to the quantization step, which counteracts the constraining effects of the attention mechanism. Table \ref{t2} shows the performances of attention-based models on the FI2010 dataset, with and without the adaptive scaling step proposed in \cite{passalis2019temporal}. In most cases, the adaptive scaling step slightly degrades the performances of the attention-based models. As we will see in the next subsection, this effect is more noticeable in audio datasets. 

\begin{table}[t!]
	\begin{center}
		\caption{Performances (averaged F1 in \%) of attention-based models on the FI2010 dataset, with and without the adaptive scaling step proposed in \cite{passalis2019temporal}. The best results in each row are highlighted in bold-face}\label{t2}
		\resizebox{0.9\linewidth}{!}{
			\begin{tabular}{|c|c|c|}
				\multicolumn{3}{c}{} \\ \hline
				\textbf{Models}        & \textbf{adaptive scale}  & \textbf{no adaptive scale}               \\ \hline \hline
				\multicolumn{3}{|c|}{\textit{Prediction Horizon $H=10$}} \\ \hline \hline
				NBoF-CA 	& $66.92\scriptstyle\pm 00.08$   & $\mathbf{67.56}\scriptstyle\pm 00.02$ \\ \hline
				NBoF-TA 	& $67.34\scriptstyle\pm 00.14$   & $\mathbf{67.98}\scriptstyle\pm 00.09$ \\ \hline \hline
				TNBoF-CA 	& $\mathbf{67.84}\scriptstyle\pm 00.16$ & $67.76\scriptstyle\pm 00.05$ \\ \hline
				TNBoF-TA	& $67.16\scriptstyle\pm 00.32$  & $\mathbf{67.88}\scriptstyle\pm 00.13$ \\ \hline \hline
				\multicolumn{3}{|c|}{\textit{Prediction Horizon $H=20$}} \\ \hline \hline
				NBoF-CA 	& $59.25\scriptstyle\pm 00.14$	 & $\mathbf{59.31}\scriptstyle\pm 00.44$ \\ \hline
				NBoF-TA    & $59.26\scriptstyle\pm 00.18$ & $\mathbf{60.10}\scriptstyle\pm 00.03$ \\ \hline \hline
				TNBoF-CA 	& $\mathbf{59.75}\scriptstyle\pm 00.35$	 & $59.73\scriptstyle\pm 00.19$ \\ \hline
				TNBoF-TA	& $59.78\scriptstyle\pm 00.11$  & $\mathbf{60.04}\scriptstyle\pm 00.24$ \\ \hline \hline
				\multicolumn{3}{|c|}{\textit{Prediction Horizon $H=50$}} \\ \hline \hline 
				NBoF-CA	  & $71.93\scriptstyle\pm 00.14$ & $\mathbf{73.25}\scriptstyle\pm 00.27$ \\ \hline
				NBoF-TA   & $47.89\scriptstyle\pm 23.87$ & $\mathbf{73.02}\scriptstyle\pm 00.04$ \\ \hline \hline
				TNBoF-CA  & $71.30\scriptstyle\pm 00.29$  & $\mathbf{73.77}\scriptstyle\pm 00.37$ \\ \hline
				TNBoF-TA  & $72.32\scriptstyle\pm 00.05$ & $\mathbf{73.40}\scriptstyle\pm 00.08$ \\ \hline 
				
			\end{tabular}
		}
	\end{center}
\end{table}

\subsection{Audio Analysis Experiments}

\begin{table}[t!]
	\begin{center}
		\caption{Audio analysis results of FMA and TUT-UAS2018 datasets. The second column shows performances (test accuracy in \%) of all models without using any convolution layer as preprocessing layers. The third column shows the corresponding performances (test accuracy in \%)  when additional convolution layers were used as preprocessing layers. The best results in each column are highlighted in bold-face.}\label{t3}
		\resizebox{0.83\linewidth}{!}{
			\begin{tabular}{|c|c|c|}
				\multicolumn{3}{c}{} \\ \hline
				\textbf{Models}        & \textbf{without conv}  & \textbf{with conv}               \\ \hline \hline
				\multicolumn{3}{|c|}{\textbf{FMA Dataset}} \\ \hline \hline
				
				GRU \cite{cho2014learning}	& $33.87\scriptstyle\pm 00.27$ & $42.06\scriptstyle\pm 00.92$    \\ \hline \hline 
				NBoF \cite{passalis2019temporal}	& $35.65\scriptstyle\pm 01.41$ & $38.83\scriptstyle\pm 01.83$ \\ \hline
				NBoF-CA (our) & $\mathbf{38.29}\scriptstyle\pm 00.95$ & $41.08\scriptstyle\pm 01.85$	 \\ \hline
				NBoF-TA (our) & $36.79\scriptstyle\pm 00.29$ & $41.46\scriptstyle\pm 01.64$	 \\ \hline \hline
				TNBoF \cite{passalis2018temporal}	& $35.25\scriptstyle\pm 03.50$ & $39.13\scriptstyle\pm 00.53$ \\ \hline 
				TNBoF-CA (our)	& $37.29\scriptstyle\pm 00.66$ & $\mathbf{42.67}\scriptstyle\pm 01.23$  \\ \hline
				TNBoF-TA (our)	& $36.50\scriptstyle\pm 01.49$ & $42.58\scriptstyle\pm 00.91$  \\ \hline \hline
				\multicolumn{3}{|c|}{\textbf{TUT-UAS2018 Dataset}} \\ \hline \hline
				GRU \cite{cho2014learning}	& $56.83\scriptstyle\pm 00.78$  & $56.89\scriptstyle\pm 00.93$ \\ \hline \hline 
				NBoF \cite{passalis2019temporal}	& $52.02\scriptstyle\pm 00.18$  & $55.92\scriptstyle\pm 01.40$  \\ \hline
				NBoF-CA (our)	& $\mathbf{56.89}\scriptstyle\pm 00.17$ & $\mathbf{57.68}\scriptstyle\pm 00.65$	 \\ \hline
				NBoF-TA (our)   & $56.09\scriptstyle\pm 00.25$ & $57.63\scriptstyle\pm 00.30$   \\ \hline \hline
				TNBoF \cite{passalis2018temporal}	& $52.62\scriptstyle\pm 00.78$  & $55.30\scriptstyle\pm 00.13$ \\ \hline
				TNBoF-CA (our)	& $56.19\scriptstyle\pm 00.23$ & $56.73\scriptstyle\pm 00.51$	 \\ \hline
				TNBoF-TA (our)	& $56.34\scriptstyle\pm 00.62$ & $57.33\scriptstyle\pm 00.20$  \\ \hline
				
			\end{tabular}
		}
	\end{center}
\end{table}

One of the important types of sequence data is audio recordings. In this subsection, we present our empirical analysis using two audio datasets, representing two different applications in audio signal analysis: music genre recognition and acoustic scene classification. 

In the first application, the objective is to train an acoustic system that recognizes the genre of a short musical recording. For this purpose, we conducted experiments using the \textit{small subset} of the FMA dataset \cite{defferrard2016fma}, which contains $8000$ tracks coming from the $8$ most popular genres: \textit{pop, instrumental, experimental, folk, rock, international, electronic}, and \textit{hip-hop}. Each audio clip is $30$s long, which is transformed to Mel-spectrogram representation with $128$ frequency bands using a window of $10$ms with an overlap of $2.5$ms. The preprocessing step results in the input sequence having dimensions of $128\times 640$. 

In the second application, the objective is to train an acoustic system that can classify the type of environment based on its surrounding sounds. For this application, we used the TUT-UAS2018 dataset \cite{tut-uas2018}, which contains $8640$ audio clips recorded from $10$ urban acoustic scenes: \textit{airport, shopping\_mall, metro\_station, street\_pedestrian, public\_square, street\_traffic, tram, bus, metro, park}. Similar to the FMA dataset, we also transformed each audio clip to Mel-spectrogram with $128$ frequency bands using a window of $40$ms with an overlap of $20$ms, which results in the input sequence of size $128\times 500$.  

For both applications, we report the test accuracy as the performance metric. Experiment results on FMA and TUT-UAS2018 dataset are shown in Table \ref{t3}. Performances of all models with and without using convolution layers for feature extraction are presented in the second and third columns, respectively. 

\begin{table}[t!]
	\begin{center}
		\caption{Performances (test accuracy in \%) of attention-based models on FMA and TUT-UAS2018 datasets, with and without the adaptive scaling step proposed in \cite{passalis2019temporal}. The best results in each row are highlighted in bold-face.}\label{t4}
		\resizebox{0.9\linewidth}{!}{
			\begin{tabular}{|c|c|c|}
				\multicolumn{3}{c}{} \\ \hline
				\textbf{Models}        & \textbf{adaptive scale}  & \textbf{no adaptive scale}               \\ \hline \hline
				\multicolumn{3}{|c|}{\textbf{FMA Dataset}} \\ \hline \hline
				NBoF-CA  & $38.37\scriptstyle\pm 02.04$ & $\mathbf{41.08}\scriptstyle\pm 01.85$	 \\ \hline
				NBoF-TA  & $34.29\scriptstyle\pm 05.42$ & $\mathbf{41.46}\scriptstyle\pm 01.64$	 \\ \hline \hline
				TNBoF-CA	& $39.96\scriptstyle\pm 00.66$ & $\mathbf{42.67}\scriptstyle\pm 01.23$  \\ \hline
				TNBoF-TA	& $40.21\scriptstyle\pm 00.16$ & $\mathbf{42.58}\scriptstyle\pm 00.91$  \\ \hline \hline
				\multicolumn{3}{|c|}{\textbf{TUT-UAS2018 Dataset}} \\ \hline \hline
				NBoF-CA 	& $40.57\scriptstyle\pm 14.36$ & $\mathbf{57.68}\scriptstyle\pm 00.65$	 \\ \hline
				NBoF-TA   & $56.62\scriptstyle\pm 00.39$  & $\mathbf{57.63}\scriptstyle\pm 00.30$   \\ \hline \hline
				TNBoF-CA	& $56.59\scriptstyle\pm 00.51$  & $\mathbf{56.73}\scriptstyle\pm 00.51$	 \\ \hline
				TNBoF-TA	& $55.05\scriptstyle\pm 01.25$ & $\mathbf{57.33}\scriptstyle\pm 00.20$  \\ \hline
				
			\end{tabular}
		}
	\end{center}
\end{table}

In the FMA dataset, we can easily observe significant improvements in all models when using additional convolution layers. Without any convolution layer, the NBoF and TNBoF models outperform the GRU model on average, however, with larger variances. The order reverses when additional convolution layers were used: the GRU model enjoys a huge benefit from the preprocessing layers, outperforming the NBoF and TNBoF models. In both scenarios, i.e., with or without convolution layers, the proposed attention block greatly enhances the baseline NBoF and TNBoF models, making them the best performing models in this task. 

\begin{table}[t!]
	\begin{center}
		\caption{Audio analysis results under noisy data setting. No preprocessing convolution layer was used in this setting.}\label{t5}
		\resizebox{0.61\linewidth}{!}{
			\begin{tabular}{|c|c|c|}
				\multicolumn{2}{c}{} \\ \hline
				\textbf{Models}       & \textbf{test accuracy}               \\ \hline \hline
				\multicolumn{2}{|c|}{\textbf{Noisy FMA Dataset}} \\ \hline \hline
				
				GRU \cite{cho2014learning}	& $31.04\scriptstyle\pm 01.43$  \\ \hline \hline 
				NBoF \cite{passalis2019temporal}	& $31.54\scriptstyle\pm 00.21$ \\ \hline
				NBoF-IA (our) 	& $36.21\scriptstyle\pm 01.93$ \\ \hline \hline
				TNBoF \cite{passalis2018temporal}	& $30.71\scriptstyle\pm 00.87$ \\ \hline
				TNBoF-IA (our)	& $\mathbf{36.67}\scriptstyle\pm 00.95$  \\ \hline \hline
				\multicolumn{2}{|c|}{\textbf{Noisy TUT-UAS2018 Dataset}} \\ \hline \hline
				GRU \cite{cho2014learning}	& $56.17\scriptstyle\pm 01.31$ \\ \hline \hline 
				NBoF \cite{passalis2019temporal}	& $41.73\scriptstyle\pm 12.66$  \\ \hline 
				NBoF-IA (our)	& $\mathbf{56.79}\scriptstyle\pm 00.86$	 \\ \hline \hline 
				TNBoF \cite{passalis2018temporal}	& $51.48\scriptstyle\pm 00.85$ \\ \hline 
				TNBoF-IA (our)	& $56.04\scriptstyle\pm 00.42$  \\ \hline
				
			\end{tabular}
		}
	\end{center}
\end{table}

In the TUT-UAS2018 dataset, while adding convolution layers leads to noticeable improvements for the baseline NBoF and TNBoF, we observe no similar improvement for the GRU model. Similar to the FMA dataset, we observe consistent performance boost in the TUT-UAS2018 dataset by incorporating the proposed attention block to the NBoF and TNBoF models.  

Similar to Section \ref{finance}, we also conducted experiments in FMA and TUT-UAS2018 datasets to analyze the effects of the adaptive scaling step proposed in \cite{passalis2019temporal}. The results are shown in Table \ref{t4}. The results obtained from both audio analysis tasks in Table \ref{t4} are consistent with what we observe from the stock movement prediction task in Table \ref{t2}: although the adaptive scaling step can enhance NBoF and TNBoF models as demonstrated in \cite{passalis2019temporal}, the additional degrees of freedom introduced by this step negates the competition effects enforced by the attention mechanism, leading to performance degradation when combining both methods. 

In order to evaluate how well the proposed input attention mechanism (NBoF-IA, TNBoF-IA) tackles noisy data, we simulated contaminated audio data by adding $10$ synthetic frequency bands, which are generated by adding white noise to the averaged Mel coefficients. Here we should note that in this set of experiments, we did not use any convolution layers in order to gauge how well the layers of interests are resilient to noise. The results are shown in Table \ref{t5}. As can be seen from Table \ref{t5}, when moving from the noiseless to the noisy version of FMA and TUT-UAS2018 datasets, the accuracy of NBoF and TNBoF models dropped significantly. GRU models also exhibited similar behaviors, although the performance drops are less significant as compared to the NBoF and TNBoF. By incorporating the proposed input attention block to the NBoF and TNBoF models, we were able to achieve very similar performances compared to the noiseless scenario. 

\subsection{Medical Diagnosis Experiments}

\begin{table}[t!]
	\begin{center}
		\caption{Performance (averaged F1) on AF dataset}\label{t6}
		\resizebox{0.56\linewidth}{!}{
			\begin{tabular}{|c|c|c|}
				\multicolumn{2}{c}{} \\ \hline
				\textbf{Models}       & \textbf{F1}               \\ \hline \hline
				GRU \cite{cho2014learning}	& $76.42\scriptstyle\pm 00.86$  \\ \hline \hline 
				NBoF \cite{passalis2019temporal}	& $78.15\scriptstyle\pm 00.82$ \\ \hline				
				NBoF-CA (our) 	& $78.73\scriptstyle\pm 00.71$ \\ \hline
				NBoF-TA (our)	& $78.55\scriptstyle\pm 00.90$  \\ \hline \hline
				TNBoF \cite{passalis2018temporal}	& $78.27\scriptstyle\pm 01.02$ \\ \hline 
				TNBoF-CA (our) 	& $78.71\scriptstyle\pm 00.92$ \\ \hline
				TNBoF-TA (our)	& $\mathbf{79.52}\scriptstyle\pm 00.81$  \\ \hline 
				
			\end{tabular}
		}
	\end{center}
\end{table}

Medical diagnosis, which plays a crucial role in ensuring human prosperity, is inherently an intricate process. The quality of the diagnosis is highly dependent on the expertise of the examiner. Since it takes several years and a great amount of resources to train human experts, medical diagnosis tools have been actively developed over the past decades to assist human examiners. In our empirical analysis using medical data, we investigated the effectiveness of the proposed models in diagnosing cardiovascular diseases using publicly available Electrocardiogram (ECG) and Phonocardiogram (PCG) signals. 

The AF dataset focuses on the problem of atrial fibrillation detection from ECG recordings, which are provided as the development data (training set) in the Physionet/Computing in Cardiology Challenge 2017 \cite{clifford2017af}. The dataset contains $8528$ single-lead ECG recordings lasting from $9$ to $60$ seconds. The objective of the challenge was to classify a given recording into one of the $4$ classes: normal sinus rhythm, atrial fibrillation, alternative rhythm, and noise. We followed an experimental setup similar to \cite{andreotti2017comparing}, which evaluates a given model using 5-fold cross-validation. Additionally, the recordings were clipped or padded so that they have a constant length of $30$ seconds. Since a single lead ECG recording is only a univariate sequence, it is necessary to use convolution layers as preprocessing layers to extract higher-level features, before the NBoF or GRU layers. To tackle the imbalanced nature of the training set, we scaled the loss term associated with each class, with the factor inversely proportional to the number of samples in that class. 

PCG signal is often used in ambulatory diagnosis in order to evaluate the heart hemodynamic status and detect potential cardiovascular problems. The data used in our experiments come from the training set provided in the Physionet/Computing in Cardiology Challenge 2016 \cite{clifford2016classification}. The objective of the challenge is to develop an automatic classification method for the anomaly (normal versus abnormal) and quality (good versus bad) detection given a PCG recording. 

Since the length of the recordings varies greatly, from $5$ to $120$ seconds, we generated $5$s segments from the recordings for training the models; during the test phase, the models were used to classify $5$s sub-segments (with $4$s overlap) of a given recording, and the overall label is inferred from the averaged classification of the sub-segments. PCG signal captures the acoustic nature of the heart sound; thus, we extracted Mel-spectrogram with $24$ frequency bands, using a window of $25$ms with an overlap of $10$ms to represent each segment. With a smaller size compared to the AF dataset, we only employed a 3-fold cross-validation protocol for this problem. Further details regarding our experimental setup in AF and PCG datasets are provided in the Appendix.

\begin{table}[t!]
	\begin{center}
		\caption{Performance (mean of sensitivity and specificity) on PCG dataset. The higher, the better.}\label{t7}
		\resizebox{\linewidth}{!}{
			\begin{tabular}{|c|c|c|}
				\multicolumn{3}{c}{} \\ \hline
				\textbf{Models}        & \textbf{Anomaly Detection}  & \textbf{Quality Detection}               \\ \hline \hline
				
				GRU \cite{cho2014learning}	& $\mathbf{90.08}\scriptstyle\pm 00.68$ & $\mathbf{72.74}\scriptstyle\pm 01.40$    \\ \hline \hline

				NBoF \cite{passalis2019temporal}	& $50.31\scriptstyle\pm 00.42$ & $49.69\scriptstyle\pm 00.34$ \\ \hline
				NBoF-CA (our) 	& $88.09\scriptstyle\pm 00.25$ & $71.98\scriptstyle\pm 03.00$ \\ \hline
				NBoF-TA (our) 	& $89.32\scriptstyle\pm 01.02$ & $72.57\scriptstyle\pm 02.20$  \\ \hline \hline
				TNBoF \cite{passalis2018temporal}	& $54.12\scriptstyle\pm 05.18$ & $53.40\scriptstyle\pm 04.21$ \\ \hline 
				TNBoF-CA (our)	& $88.68\scriptstyle\pm 00.95$ & $72.34\scriptstyle\pm 01.32$ \\ \hline
				TNBoF-TA (our)	& $88.81\scriptstyle\pm 01.07$ & $69.45\scriptstyle\pm 00.89$  \\ \hline
				
			\end{tabular}
		}
	\end{center}
\end{table}

Table \ref{t6} shows the averaged F1 score, a metric adopted by the database \cite{clifford2017af}, of all models on the AF dataset. In Table \ref{t7}, we show the anomaly and quality detection performance. The performance metric used by the database \cite{clifford2016classification} is calculated as the mean of sensitivity and specificity scores. In the AF dataset, the averaged F1 scores obtained from the baseline NBoF and TNBoF models are significantly higher than the one obtained from the recurrent model. Although the improvement margins for the NBoF model are minor in Table \ref{t6}, both the NBoF and TNBoF models enjoy increases in performance when using the proposed attention blocks. The consistent performance gain produced by the attention blocks can also be observed in the PCG dataset in Table \ref{t7}. In this dataset, while the NBoF and TNBoF models score far below the GRU model, the attention-based models perform nearly as well as the recurrent model. 

\section{Conclusions}\label{conclusions}

In this paper, we proposed 2D-Attention, a generic attention mechanism for data represented in the form of matrices. The proposed attention computation can be used in a plug-and-play manner, and can be updated jointly with other components in a computation graph. Using the proposed attention block, we further proposed three variants of the Neural Bag-of-Features model when learning with sequence data. Our extensive experiments in financial forecasting, audio analysis and medical diagnosis demonstrated that the proposed attention consistently led to performance gains for the Neural Bag-of-Features models. Since 2D-Attention is a generic attention computation method for matrices, investigating its efficacy in other neural network models is an interesting research direction in the future works.   

\section{Acknowledgement}
This project has received funding from the European Union's Horizon 2020 research and innovation programme under grant agreement No 871449 (OpenDR). This publication reflects the authors’ views only. The European Commission is not responsible for any use that may be made of the information it contains.

\appendix

In all of our experiments, we used ADAM optimizer for stochastic optimization. Weight decay ($0.0001$) or max-norm constraint ($4.0$) was used to for regularization. In addition, dropout ($0.2$) was applied to the output of the layer before the classification layer. In all models, before the output layer, there is a fully-connected layer with $512$ neurons. For NBoF, TNBoF and the attention models, we used $256$ codewords in the quantization layer. Correspondingly, the number of units in GRU model was set to $256$. Details that are specific to each experiment are provided below:

\begin{itemize}
\item \textit{Financial Forecasting Experiments}: All models were trained for $80$ epochs, with the initial learning rate set to $0.001$. The learning rate was decreased by a factor of $0.1$ at epoch $11$ and $51$. We followed \cite{tran2018temporal} and scaled the loss term associated with each class with a factor that is inversely proportional to the number of samples of each class to counter the effect of class imbalanced. In experiments that used convolution layers as preprocessing layers, we used two 1D convolution layers, each of which has $64$ filters with the filter size set to $5$ and the stride set to $1$. Batch normalization was used after each convolution layer, followed by the ReLU activation. 

\item \textit{Audio Analysis Experiments}: The setup is similar to the financial forecasting experiments, except for the configuration of convolution layers: four 1D convolution layers with the filter size of $5$ were used; the first two convolution layers have $32$ filters, which are followed by a max-pooling layer to reduce the temporal dimension by half. The last two convolution layers have $64$ filters. After each convolution layer, we applied batch normalization, followed by ReLU activation. 

\item \textit{Medical Diagnosis Experiments}: in both AF and PCG datasets, all models were trained for $90$ epochs, with the initial learning rate set to $0.001$, which was decreased to $0.0001$ at epoch $11$, then to $0.00001$ at epoch $71$. For the AF dataset, we adopted the convolution architecture proposed in \cite{andreotti2017comparing} as the first computation block in all models. For PCG dataset, we used five 1D convolution layers with the filter size set to $3$ as the preprocessing layers: the first two layers have $32$ filters with strides of $1$; the third layer has $64$ filters with strides of $2$; the fourth layer has $64$ filters with strides of $1$; the last layer has $128$ filters with strides of $2$. After each convolution layer, we applied batch normalization, followed by ReLU activation.
\end{itemize}

\bibliography{reference}
\bibliographystyle{ieeetr}

\end{document}